\documentclass{article}
\usepackage[utf8]{inputenc}
\usepackage{graphicx}
\usepackage{subfig}
\usepackage{authblk}
\usepackage[compress]{natbib}
\usepackage[dvipsnames]{xcolor}
\usepackage[colorlinks=true,urlcolor=TealBlue,citecolor=Bittersweet,linkcolor=black]{hyperref}
\usepackage[left=3.2cm,right=3.2cm,top=3cm,bottom=3.5cm]{geometry}

\title{\texttt{normflows}: A PyTorch Package for Normalizing Flows}

\author[1,2,@]{\normalsize \bfseries Vincent Stimper}
\author[1]{\normalsize \bfseries David Liu} 
\author[1]{\normalsize \bfseries Andrew Campbell}
\author[2]{\normalsize \bfseries Vincent Berenz}
\author[1]{\normalsize \bfseries Lukas Ryll}
\author[2]{\normalsize \bfseries Bernhard Schölkopf}
\author[1]{\normalsize \bfseries José Miguel Hernández-Lobato} 

\affil[1]{\small University of Cambridge, Cambridge, United Kingdom} 
\affil[2]{\small Max Planck Institute for Intelligent Systems, Tübingen, Germany}
\affil[@]{Corresponding author: \texttt{vstimper@tue.mpg.de}}

\date{\vspace*{-2em}}

\makeatletter
\newcommand\footnoteref[1]{\protected@xdef\@thefnmark{\ref{#1}}\@footnotemark}
\makeatother

\begin{document}

\maketitle

\section*{Summary}

Normalizing flows model probability distributions through an expressive tractable density \citep{Tabak2010,tabak2013family,rezende2015variational}. They transform a simple base distribution, such as a Gaussian, through a sequence of invertible functions, which are referred to as layers. These layers typically use neural networks to become very expressive. Flows are ubiquitous in machine learning and have been applied to image generation \citep{kingma2018glow,grcic2021}, text modeling \citep{wang2019text}, variational inference \citep{rezende2015variational}, approximating Boltzmann distributions \citep{noe2019boltzmann}, and many other problems \citep{kobyzev2021,papamakarios2021normalizing}. 

Here, we present \texttt{normflows}, a Python package for normalizing flows. It allows to build normalizing flow models from a suite of base distributions, flow layers, and neural networks. The package is implemented in the popular deep learning framework PyTorch \citep{paszke2019pytorch}, which simplifies the integration of flows in larger machine learning models or pipelines. It supports most of the common normalizing flow architectures, such as Real NVP \citep{dinh2017RealNVP}, Glow \citep{kingma2018glow}, Masked Autoregressive Flows \citep{papamakarios2017}, Neural Spline Flows \citep{muller2019neural,durkan2019neuralspline}, Residual Flows \citep{chen2019residual}, and many more. The package can be easily installed via \texttt{pip} and the code is publicly available on GitHub\footnote{\label{fn:repo}\url{https://github.com/VincentStimper/normalizing-flows}}.

\section*{Statement of need}

\texttt{normflows} focuses on flows that are composed of discrete transformations, as opposed to continuous normalizing flows \citep{Chen2018a,papamakarios2021normalizing}. There are several other packages implementing discrete normalizing flows, such as TensorFlow Probability \citep{dillon2017} for TensorFlow, \texttt{distrax} \citep{deepmind2020jax} for JAX, and \texttt{nflows} \citep{nflows} and \texttt{FrEIA} \citep{freia} for PyTorch. However, none of them support the two popular flow architectures, residual and autoregressive flows, within a single package, while we do so.

Moreover, \texttt{normflows} stands out by providing tools that are often used when approximating Boltzmann distributions. First, sampling layers needed for Stochastic Normalizing Flows \citep{wu2020stochasticNF,nielsen2020} are included. Second, Neural Spline Flows on circular coordinates are supported \citep{rezende2020}, which can be combined with standard coordinates on bounded or unbounded intervals. They are needed when modeling the internal coordinates of molecules consisting of angles and lengths \citep{Midgley2022}. Furthermore, there is an extension for \texttt{normflows} that adds Boltzmann distributions as targets as well as flow layers converting between Cartesian and internal coordinates \citep{boltzgen}.

Our package has already been used in several scientific projects and publications \citep{campbell2021gradient,stimper2021,Midgley2022}. Due to its modular nature, \texttt{normflows} can be easily extended to house new flow layers, base distributions, or other tools. For instance, \cite{stimper2021} extends the package by adding resampled base distributions, which overcome an architectural weakness of normalizing flows and make them more expressive.

\section*{Examples}

\begin{figure}
    \centering
    \subfloat[]{\includegraphics[width=0.57\textwidth]{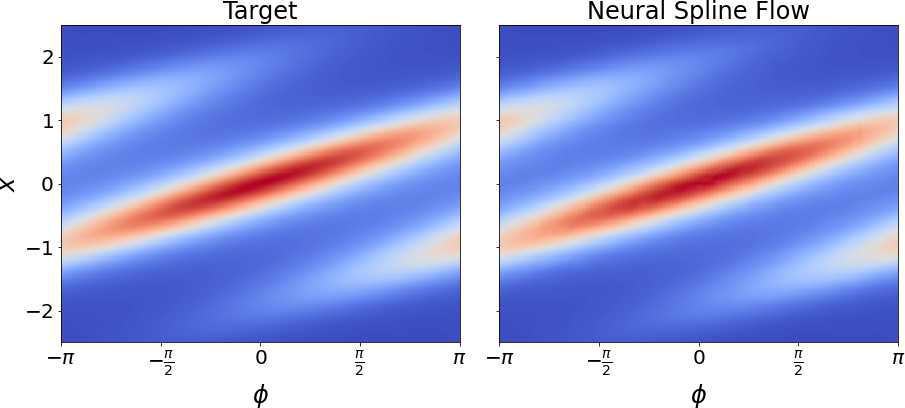}}
    \hfill
    \subfloat[]{\raisebox{0.6cm}{\includegraphics[width=0.38\textwidth]{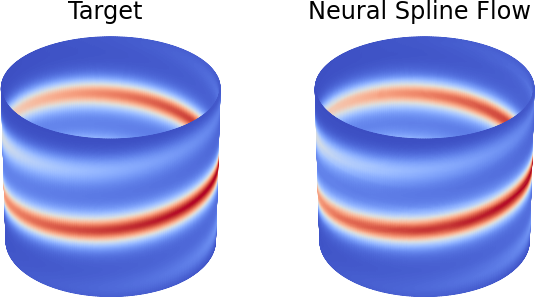}}}
    \caption{Target density defined on a cylinder surface, having an unbounded coordinate $x$ and a circular coordinate $\phi$. A Neural Spline Flow is fit to it, being almost indistinguishable from the target. (a) shows the densities in 2D and (b) is a visualization on the cylinder surface.}
    \label{fig:example}
\end{figure}

In the GitHub repository of our package\footnoteref{fn:repo}, we provide various examples illustrating how to use it. We show how to build a flow model from a base distribution, a list of flow layers, and, optionally, a target distribution. They can be trained by computing a loss through the respective methods provided and minimizing it with the standard PyTorch optimizers. We show how to approximate simple 2D distributions, but, moreover, apply flows to images through the multiscale architecture \citep{dinh2017RealNVP}, which \texttt{normflows} provides as well. Furthermore, there is an example of how to build a variational autoencoder with normalizing flows as well.

Here, we want to illustrate a strength of \texttt{normflows}, i.e. that it can deal with combinations of standard and circular coordinates. Therefore, we consider a distribution of two random variables, $x$ and $\phi$. $x$ follows a Gaussian distribution with density $p(x) = \mathcal{N}(x|0, 1)$ and $\phi$ has a circular von Mises distribution such that $p(\phi|x) = \mathcal{M}(\phi|\mu(x), 1)$ with $\mu(x) = 3x$. We train a Neural Spline Flow with an unbound and a circular coordinate to approximate the target distribution $p(x, \phi) = p(x) p(\phi|x)$ by minimizing the reverse Kullback-Leibler divergence. As shown in \autoref{fig:example}, the density of the flow is almost indistinguishable from the target.

\section*{Acknowledgements}

We thank Laurence Midgley and Timothy Gebhard for their valuable contributions to the package.
Moreover, we thank everyone who contacted us via mail or on GitHub for the valuable feedback and spotting bugs.

Jos\'e Miguel Hern\'andez-Lobato acknowledges support from a Turing AI Fellowship under grant EP/V023756/1.
This work was supported by the German Federal Ministry of Education and Research (BMBF): Tübingen AI Center, FKZ: 01IS18039B; and by the Machine Learning Cluster of Excellence, EXC number 2064/1 - Project number 390727645.

\bibliographystyle{plainnat}
\bibliography{references}

\end{document}